\documentclass[12pt, twoside, a4paper]{article}
\usepackage{imm17_4}

\pdfoutput=1

\usepackage{tabularx}
\usepackage{multirow}
\usepackage{pbox}
\usepackage{rotating}

\usepackage{color, colortbl}
\hypersetup{pdftitle=Deep Learning Algorithms for Signal Recognition in Long Perimeter Monitoring Distributed Fiber Optic Sensors}
\hypersetup{pdfauthor={A. V. Makarenko, Constructive Cybernetics Research Group}}
\hypersetup{pdfsubject={Deep Learning}}
\hypersetup{pdfkeywords={deep learning, fiber optic vibration sensors, signal recognition, events classification, t-SNE visualization}}

\newcommand{\MaincltStr}{
arXiv.org (Proceedings of the IEEE International Workshop MLSP~2016)}

\newcommand{\Leftclt}{A.\,V.\,Makarenko}

\newcommand{\Rightclt}{Deep Learning Algorithms for Signal Recognition\ldots}

\newcommand{\UDK}{PACS: 07.05.Mh, 42.81.Wg, 43.60.Lq, 43.60.Mn, 43.60.Np, 43.60.Rw.}

\begin{document}

\Mainclt 

\begin{center}
\Large{\bf Deep Learning Algorithms for Signal Recognition \\in Long Perimeter Monitoring Distributed \\Fiber Optic Sensors}\\[2ex]
\end{center}

\begin{center}
\large\bf{A.\,V.\,Makarenko}\supit{a,}\supit{b,}
\footnote{E-mail: avm.science@mail.ru}\\[2ex]
\end{center}

\begin{center}
\supit{a}\normalsize{Constructive Cybernetics Research Group}
\\
\normalsize{P.O.Box~560, Moscow, 101000 Russia}\\[3ex]

\supit{b}
\normalsize{Institute of Control Sciences, Russian Academy of Sciences}
\\
\normalsize{ul.~Profsoyuznaya~65, Moscow, 117977 Russia}\\[3ex]
\end{center}

{\small Received April 27, 2016;\;in final form, July 28, 2016.}
\begin{quote}\small
{\bf Abstract}. In this paper, we show an approach to build deep learning algorithms for recognizing signals in distributed fiber optic monitoring and security systems for long perimeters. Synthesizing such detection algorithms poses a non-trivial research and development challenge, because these systems face stringent error (type I and II) requirements and operate in difficult signal-jamming environments, with intensive signal-like jamming and a variety of changing possible signal portraits of possible recognized events. To address these issues, we have developed a two-level event detection architecture, where the primary classifier is based on an ensemble of deep convolutional networks, can recognize 7 classes of signals and receives time-space data frames as input. Using real-life data, we have shown that the applied methods result in efficient and robust multiclass detection algorithms that have a high degree of adaptability.
\end{quote}

\begin{Keyworden}
deep learning, fiber optic vibration sensors, signal recognition, events classification, t-SNE visualization.
\end{Keyworden}


\setcounter{equation}{0}
\setcounter{lem}{0}
\setcounter{teo}{0}



\section{Introduction}
\label{sec:intro}

Today, distributed fiber optic vibration sensors are used in many areas and undergo rapid development, improving their sensitivity and selectivity characteristics~\cite{bib:article_Masoudi_RevSciInstr_2016_87_011501}. Due to a number of design advantages (feasibility, full passivity, negligible response to electromagnetic radiation, compatibility with off-the-shelf telecommunication cables, etc.), fiber optic sensors are very common in monitoring and security of long perimeters, one of the key applications being monitoring of major pipelines.

\begin{figure}[htb]
\begin{minipage}[b]{1.0\linewidth}
\centerline{\includegraphics[width=8.4cm]{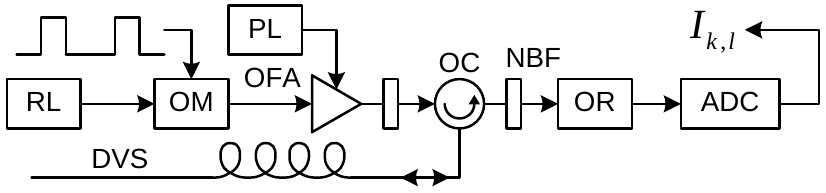}}
\end{minipage}
\caption{Schematic structural and functional diagram of the coherence optical time domain reflectometer: RL -- reference laser; OM -- electrooptical light modulator; OFA -- optical fiber amplifier; PL -- pump laser; NBF -- narrow-band filter; OC -- optical circulator; DVS -- distributed optical fibre vibration sensor; OR -- optical receiver; ADC -- analog-to-digital converter; $I_{k,\,l}$ -- intensity of the signal registered from the $l$th spatial element of the sensor at moment~$k$.}
\label{fig:Gen_System_Scheme}
\end{figure}

One of the types of fiber optic systems operates based on time-domain detection of backward Rayleigh scattered light of short laser pulse injected into the cable~\cite{bib:article_Masoudi_RevSciInstr_2016_87_011501}. Coherence reflectometer uses a narrowband high-stable reference laser as the light source (see Fig.~\ref{fig:Gen_System_Scheme} for the schematic structural and functional diagram). This allows to sum the intensities of the backscattered signals with phase delay factored in. Meanwhile, vibrations in the environment, where the sensor operates, cause micro-vibrations in the optic fiber of the sensor (shift amplitudes from 50 nm), displacing Rayleigh scattering centers relative to each other. As a result, spatially localized disturbances of the sensor cause local modulation of coherence reflectogram intensity.

Analyzing time-frequency and spatial features of the intensity of local modulation in the coherence reflectogram allows to build algorithms for detection and classification of vibration and acoustic fields disturbances in the vicinity of monitored perimeters~\cite{bib:article_Mahmoud_PhotonicSensors_2012_3}. Note that there are complicating factors in building such algorithms. The major ones are:
\begin{itemize}
\setlength{\parindent}{0pt}
\setlength{\parskip}{0.2ex plus 0.17ex minus 0.2ex}
\item deviations in local temperature of the sensor causes ghost modulations of the reflectogram;
\item deviations in laser light phase (frequency) cause local sections of the sensor to become temporarily insensitive;
\item great variety of changing possible signal portraits of detected events requires ingenious approaches to building dictionaries for identification of events;
\item broad range of changing mechanical and acoustic properties of the environment where the sensor operates leads to significant changes in the signal portraits of recognized events.
\end{itemize}
The abovementioned factors as well as intensive signal-like jamming make for a complex signal-jamming environment that the event recognition algorithms have to process. Another complicating factor is the stringent requirements to monitoring and security systems at critical facilities in terms of type~I and~II error rates in detection and classification of events in the monitored area.

Therefore, a considerable depth of adaptability is essential to efficient and reliable operation of the recognition algorithms. Meeting these requirements is quite difficult if standard methods of algorithm synthesis are used (e.g., see~\cite{bib:book_Song_2002}). Moreover, our research has indicated that a number of machine learning algorithms~\cite{bib:article_Mahmoud_PhotonicSensors_2012_3, bib:article_Fedorov_RevSciInstrum_2016_3} are inadequate for this task as well. Hence, to solve this problem, we have built a distributed fiber optic sensor system using deep learning algorithms based on an ensemble of convolutional neural networks.

\section{The Problem Description}
\label{sec:problem}

We synthesize a signals recognition algorithm for a distributed fiber optic monitoring and security system for major pipelines (developers PetroLight Company and OMEGA Company, Moscow, Russia). Key specifications: monitoring perimeter length (one optical sensor) is 50~km; space step of the sensor is 2~m; ADC resolution is 16~bit; signal sampling rate for one spatial channel of the sensor is 1666~Hz.

We have defined the following classes of detected and classified signals (events) according to the specific purpose of the system:
\begin{itemize}
\setlength{\parindent}{0pt}
\setlength{\parskip}{0.2ex plus 0.17ex minus 0.2ex}
\item 0 -- "natural and technogenic jamming";
\item 1 -- "liquid spilling into the ground from pipeline leakage";
\item 2 -- "digging with hand tools";
\item 3 -- "digging with heavy excavation equipment";
\item 4 -- "drilling into the pipeline wall";
\item 5 -- "welding onto the pipeline wall";
\item 6 -- "grinding the pipeline wall".
\end{itemize}
Note that the signal (event) classes defined above address the whole range of control tasks in the monitored area of major pipelines and allow to detect conventional forms of illegal activity.

\begin{figure}[htb]
\begin{minipage}[b]{1.0\linewidth}
\centerline{\includegraphics[width=8.4cm]{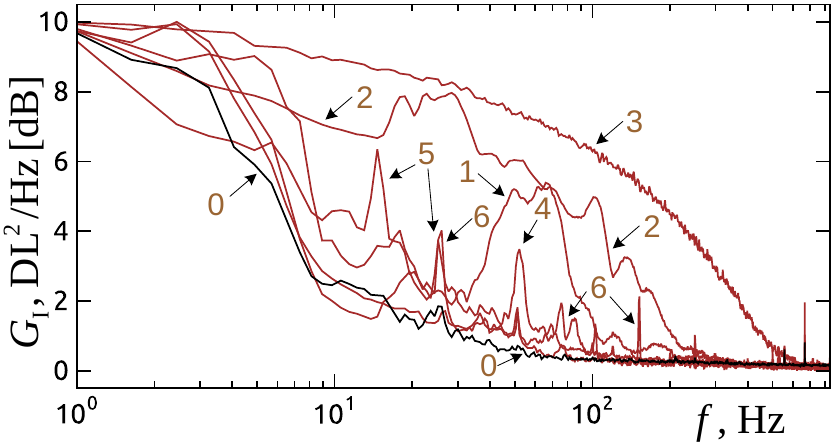}}
\end{minipage}
\caption{Power spectrums of the investigated signals averaged in time (computed using the Hamming window), DL is digital level of ADC, digits~$0,\ldots,6$ represent classes of the investigated signals.}
\label{fig:PSD_Sg_JE_FrMn}
\end{figure}

The Fig.~\ref{fig:PSD_Sg_JE_FrMn} shows power spectrums of the investigated signals averaged in time (noncoherent combining interval~$\approx 20$~min) and normalized in~$0$ to $10$~dB range.

The spectrums in Fig.~\ref{fig:PSD_Sg_JE_FrMn} are typical of the detected signals and, as can be clearly seen, considerably different from each other. In light of this, we have attempted to synthesize a classifier based on the shape of the integral spectrum using Support vector Machine, Random Forest and Fully Connected Neural Networks. Informative features were synthesized using "hand-engineered" feature extraction technique. The results have turned out below satisfactory, because the resulting classifiers lack robustness in the changing signal-noise environment. Additional investigations have indicated that, due to the factors presented in the Introduction, the shape of spectrums stabilizes (manifests) only when the generation intervals are longer. At the same time, their shape can significantly change depending on the disturbance source, leading to signal profile variation effect). In fact, the problem described above is quite common when "hand-engineered" feature extraction technique is applied~\cite{bib:report_Besaw_SPIE_2014}.

With regard to these new findings, we have decided to synthesize a detection algorithm based on convolutional neural networks, which have been proven effective for similar applications, e.g., see~\cite{bib:article_Abdel-Hamid_TransAudioSpeechLangProc_2014_22_1533, bib:report_Chen_DSAA_2014, bib:Book_Goodfellow-et-al_2016}.

To train and test the algorithms, we have built a reference dataset of real-life signals and jamming under the conditions similar to normal modes of operation of the instrument. Dataset samples have been recorded in three geographical zones, differing in ground properties, cable installation, major pipeline specifications, and natural and manmade jamming patterns. The recordings cover several calendar seasons in all three zones, with ground freezing, thawing and watering factored in. It is also worth noting that we have introduced significant variation into the key parameters of signal sources in order to obtain a wider range of signal portraits of recognized events and to enhance the generalization ability of the classifier. For example, to record class~1 signals (Leakage) we have used pipes of various diameters, various liquid pressures and defects of various sizes; for class~5 signals (Welding) we have used electrodes of various diameters and various welding current settings.

\section{The Target Data Processing Circuit}

\subsection{The Circuit Structure}
\label{sec:contour}

\begin{figure}[htb]
\begin{minipage}[b]{1.0\linewidth}
\centerline{\includegraphics[width=7.8cm]{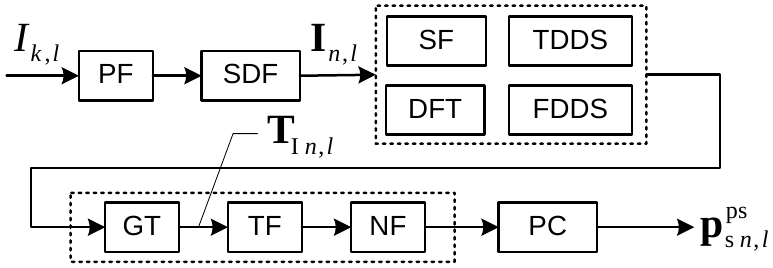}}
\end{minipage}
\caption{Schematic structural and functional diagram of the target data processing circuit: PF -- primary filter; SDF -- shaper of data frame; SF -- secondary adaptation filters; DFT -- digital Fourier transform; TDDS -- time domain decision statistics; FDDS -- frequency domain decision statistics; GT -- generation of primary features array; TF -- transformation of primary features; NF -- normalization of primary features; PC -- primary classifier.}
\label{fig:Gen_DataProc_Scheme}
\end{figure}

The Fig.~\ref{fig:Gen_DataProc_Scheme} shows the schematic structural and functional diagram of the synthesized target data processing circuit for primary signal classification.

The~ADC provides digital sample stream as raw data input to be filtered channel by channel by the primary filter in order to remove components that fall outside the valid signals range. Then, we feed the filtered data into a block that uses the digital sample stream to build single-channel time-space data frames:
\begin{equation}
\begin{aligned}
& \mathbf{I}_{n,\,l} = \{I_{k',\,l}\}^{k_e}_{k'=k_b} = \bigl[
I_{k_b,\,l},\,I_{k_b+1,\,l},\,\ldots,\,I_{k_e,\,l}\bigr],\\[3pt]
& k_b = n\, K'/f_d,\quad
k_e = k_b+K'-1,
\end{aligned}
\end{equation}
where~$n$ and~$K'$ are the number and the size of the data frame, $f_d = 1,\ldots,8$ is the overlap factor. For each data frame, we compute a decision statistics set for both the time domain (coefficients of kurtosis, skewness, etc.) and the frequency domain (filter bank, etc). Note that to improve the robustness and efficiency of the recognition algorithms we have introduced a secondary filters, which ensures the adaptability of the algorithms.

Next, we use the decision statistics set to generate a primary features array~$\mathbf{T}_{\mathrm{I}\,n,\,l}$, which has a specific structure coordinated with the primary classifier input. We transform this array applying specific rules, normalize and feed it into the primary classifier that provides an estimation of confidence (posterior probability) attributing time-space data frames to one of the signal classes~$\mathbf{p}^\mathrm{ps}_{\mathrm{s}\,n,\,l}$.

\begin{figure}[htb]
\begin{minipage}[b]{1.0\linewidth}
\centerline{\includegraphics[width=5.4cm]{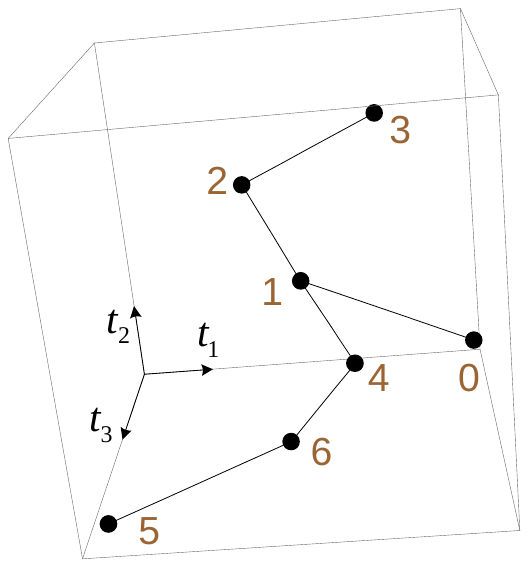}}
\end{minipage}
\caption{Median t-SNE cluster centers for the signal classes.}
\label{fig:tSNE_3D_Md_Ts_Set}
\end{figure}

At the stage of defining and analyzing primary features, we have applied nonlinear decrease dimensionality reduction t-SNE method~\cite{bib:article_Maaten_JMLR_2008_9_2579}. We have employed a portion of the test set, 35K~samples used in total, with the recognized signal classes equally distributed within the set (see Sec.~\ref{sec:learning}). We have run PCA analysis and used the first 64~principal components for the source dataset. We have used~3D t-SNE visualization. Fig.~\ref{fig:tSNE_3D_Md_Ts_Set} shows median cluster centers, which have been interconnected through minimum spanning tree algorithm, with the distances between graph node minimized.

Note that the obtained t-SNE tree of cluster centers has proved to be robust against variations in the key parameters of the nonlinear decrease dimensionality reduction algorithm. In some cases, the~0-1 connection changed for~0-4, whereas the rest of the graph remained the same. With regard to the~t-SNE algorithm properties, this stability of partition indicates that the data structure in the domain of the selected informative features possesses a high level of invariance.

The Fig.~\ref{fig:tSNE_3D_Md_Ts_Set} indicates that the 7~signal classes, that we have defined among the identified primary features (specifically, their main entries), can be clearly differentiated between. We see that the signals of classes~1 (Leakage) and~4 (Drilling) look closest to class~0 signals (Background), whereas the signals of classes~3 (Excavation) and~5 (Welding) are the most distinct from the Background, see Table~\ref{tbl:MatrixL2_TSNE}.

\begin{table}[h!]
\begin{center}
\caption{Distances between cluster centers.}
\label{tbl:MatrixL2_TSNE}
\small{
\begin{tabularx}{292.75pt}[c]{
|p{0.018\linewidth}
|p{0.07\linewidth}
|p{0.07\linewidth}
|p{0.07\linewidth}
|p{0.07\linewidth}
|p{0.07\linewidth}
|p{0.07\linewidth}|}\hline
\#&$1$&$2$&$3$&$4$&$5$&$6$\\ \hline
$0$&$0.486$&$0.782$&$1.000$&$0.551$&$0.881$&$0.626$\\ \hline
$1$&       &$0.310$&$0.575$&$0.201$&$0.534$&$0.312$\\ \hline
$2$&       &       &$0.307$&$0.353$&$0.558$&$0.440$\\ \hline
$3$&       &       &       &$0.545$&$0.715$&$0.626$\\ \hline
$4$&       &       &       &       &$0.416$&$0.153$\\ \hline
$5$&       &       &       &       &       &$0.281$\\ \hline
\end{tabularx}
}
\end{center}
\end{table}

The results essentially correspond to the shape of the integral power spectrum of signals (see Fig.~\ref{fig:PSD_Sg_JE_FrMn}). Later, the analysis of the error matrix of the trained classifier have corroborated the results as well (see Table~\ref{tbl:ConfMatrix} for example).

\subsection{The Primary Classifier}
\label{sec:primary}

\begin{figure}[htb]
\begin{minipage}[b]{1.0\linewidth}
\centerline{\includegraphics[width=7.9cm]{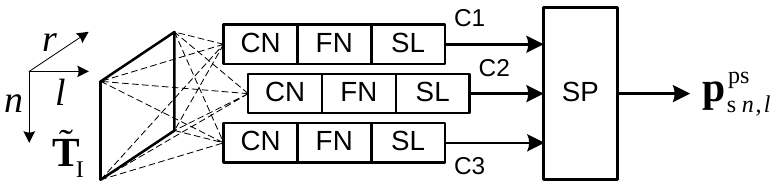}}
\end{minipage}
\caption{Schematic structural and functional diagram of the primary classifier: CN -- convolution network; FN -- fully connected network; SL -- softmax layer; SP -- aggregating rule: $r=1,\ldots,R_\mathrm{M}$, where~$R_\mathrm{M}$ is primary feature vector dimensionality.}
\label{fig:Gen_PC_Scheme}
\end{figure}

The Fig.~\ref{fig:Gen_PC_Scheme} shows the scheme of the primary classifier structure.

The Fig.~\ref{fig:Gen_PC_Scheme} shows that the primary classifier is an ensemble of three forward-propagation deep neural networks (C1--C3), where convolution and fully connected layers are consecutively combined. The key difference between the networks is their convolution kernel configuration.
Note that the use of three networks is a compromise between the quality of classifier decisions and its speed of real-time operation. We feed a~2D data blob with the input format of~$N_\mathrm{M}\times R_\mathrm{M}$ into each of the networks so that the primary classifier processes the spatial input channel by channel. The output layer of each network has~$7$ neurons (by the number of recognized signal classes) and uses SoftMax activation function:
\begin{equation}
\sigma(\mathbf{z})_i = \cfrac{\mathrm{e}^{z_i}}{\sum_{j=0}^6 \mathrm{e}^{z_j}},\quad
i= 0,\ldots,6.
\end{equation}
This allows to interpret the values of the output vector~$\sigma(\mathbf{z})$ as event probabilities that form a full group when taken together.

We can obtain partial solutions for C1--C3 networks using a threshold rule:
\begin{equation}\label{eq:ThresCl}
\begin{aligned}
&c^{\mathrm{ps}\, (j)}_{\mathrm{s}\,n,\,l}=
\begin{cases}
i^* & \sigma(\mathbf{z})^{(j)}_{i^*} \geqslant \bigl(\boldsymbol{\alpha}^{(j)}_\mathrm{c}\bigr)_{i^*},\\
0   & \text{otherwise}.
\end{cases},\\
&i^* = \arg\max_i \sigma(\mathbf{z})^{(j)}_{n,\,l}, \quad
i = 0,\ldots,6,\quad
j = 1,\ldots,3,
\end{aligned}
\end{equation}
where~$\boldsymbol{\alpha}^{(j)}_\mathrm{c}$ is the vector set for decision thresholds in~$j$th network. The most popular and, in a way, the simplest rule for generating a single decision based on partial solutions for C1--C3 networks is two-out-of-three selection. With the transitivity property factored in, it can be defined as follows:
\begin{equation}\label{eq:rule23}
\begin{aligned}
&c^\mathrm{ps}_{\mathrm{s}\,n,\,l} = \max\Bigl\{
c^{\mathrm{ps}\, (1)}_{\mathrm{s}\,n,\,l}\,d_{12},\,
c^{\mathrm{ps}\, (1)}_{\mathrm{s}\,n,\,l}\,d_{13},\,
c^{\mathrm{ps}\, (2)}_{\mathrm{s}\,n,\,l}\,d_{23}\Bigr\},\\
&d_{ij} = \delta\Bigl[
c^{\mathrm{ps}\, (i)}_{\mathrm{s}\,n,\,l}, c^{\mathrm{ps}\, (j)}_{\mathrm{s}\,n,\,l}\Bigr],\quad
i,j=1,\ldots,3,
\end{aligned}
\end{equation}
where~$\delta[\cdot,\,\cdot]$ is the Kronecker delta.

The standard approach based on rule~(\ref{eq:rule23}) has a significant drawback, as the information on posterior probability of time-space data frame attribution is lost too early. This puts a critical limitation on the system, because the receptive fields of the primary classifier are separated (isolated) into spatial channels (where the input is a data blob of~$N_\mathrm{M}\times R_\mathrm{M}$ format), meaning that the classifier cannot either select a signal source based on its space width or differentiate between a moving and a static signal source. Such a simplification is necessary due to real-time classification and hardware limitations that are usually bypassed via secondary data processing, see Sec.~\ref{sec:second}. Note that we are researching algorithms to synthesize classifier for the input format of~$N_\mathrm{M}\times L_\mathrm{M}\times R_\mathrm{M}$, where~$L_\mathrm{M} > 1$, but the requirements to~GPU performance are lower.

In light of the above, we can propose two simple rules for generating a single decision based on partial solutions for C1--C3 networks, where the information on posterior probability of time-space data frame attribution is retained. First, averaging by~$L_2$ norm:
\begin{equation}\label{eq:select_L2}
\mathbf{p}^\mathrm{ps}_{\mathrm{s}\,n,\,l} = \frac{\mathbf{s}}{\lvert\mathbf{s}\rvert}, \quad
\mathbf{s} = \sum_{j=1}^3\sigma(\mathbf{z})^{(j)}_{n,\,l},
\end{equation}
second, selecting decision with the highest confidence (in fact -- $L_\infty$ norm):
\begin{equation}
\mathbf{p}^\mathrm{ps}_{\mathrm{s}\,n,\,l} = \sigma(\mathbf{z})^{(g)}_{n,\,l}, \:
g = \arg\max_j \sigma(\mathbf{z})^{(j)}_{n,\,l},\:
j=1,\ldots,3.
\end{equation}
It is worth noting that we have investigated more complex rules for generating a single decision based on partial solutions for C1--C3 networks as well.

\subsection{The Secondary Data Processing}
\label{sec:second}

The primary classifier output is inadequate to be directly used for generating reports on the situation in the monitored area. The reason for that is twofold. First, as has been explained above, the primary classifier cannot select signal sources based in their space width and cannot identify a moving source. Second, in order to minimize the number of undetected events and to stabilize the false alarm rate, the decisions of the primary classifier need to be glued (as well as filtered) into signal-event tracks that are used to form final event reports. We resolve these issues by using a secondary classifier.

\begin{figure}[htb]
\begin{minipage}[b]{1.0\linewidth}
\centerline{\includegraphics[width=7.3cm]{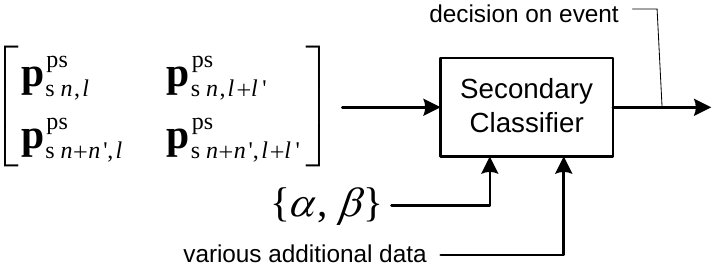}}
\end{minipage}
\caption{Schematic structural and functional diagram of the secondary classifier: $\{\alpha,\,\beta\}$  are the allowed error rates for type I and II errors.}
\label{fig:PSC_Scheme}
\end{figure}

Detailed description of the secondary data processing algorithms and synthesizing them is beyond the scope of this paper, besides we are actively researching this area at the moment. Therefore, we confine ourselves to highlighting several key issues that we have already resolved. We use vectors~$\mathbf{p}^\mathrm{ps}_{\mathrm{s}\,n,\,l}$ as the main input for the secondary classifier, thereby achieving stacking of the primary and secondary classifiers. For the secondary classifier input, we use main-data blob of~$N_\mathrm{H}\times L_\mathrm{H}\times M_\mathrm{H}$ format, where~$M_\mathrm{H}$ is determined by the number of recognized signal classes, and we set~$L_\mathrm{H} > 1$ to introduce spatial features into the algorithm (see Fig.~\ref{fig:PSC_Scheme}).
\\We build the architecture of this classifier based on deep neural networks as well. The decisions of this classifier can be controlled by the operator, who can set the allowed error rates for type I and II errors. Moreover, this classifier operates with various additional data, namely: a priory probabilities of certain event classes being detected at a given time in a specific area; primary classifier output provided by a second optical sensor (back-to-back monitoring of the perimeter by two sensors); decisions provided by other monitoring systems based on other operation principles.

\subsection{Classifier Training}
\label{sec:learning}

Prior to defining and analyzing primary features and to training and testing the classifier, we have had an operator mark up the raw dataset of signals, jammings and noises (see Sec.~\ref{sec:problem}) using automation technology. The mark-up consists of defining time-space boundaries of signals, assigning event classes and introducing additional machine-readable metadata. Next, we have separated the data into training and test sets, which do not overlap. The results of the present paper have been obtained on~487K samples in the training set and~70K samples in the test set. The shares of the event classes in the test set are uniform balanced.

We trained the three deep neural networks C1--C3 separately. We applied the single-label approach (one-hot vector) and the loss function for categorical crossentropy:
\begin{equation}
L^{(j)} = -\frac{1}{K} \sum\limits^{K}_{k=1}\sum\limits^{6}_{i=0}(\mathbf{t}_k)_i\ln\sigma(\mathbf{z})^{(j)}_{i\,k},\quad
j=1,\ldots,3,
\end{equation}
where~$K$ is the size of the training set, $\mathbf{t}_k$ is the label vector tied to the $k$th time-space data frame included into the training set. Note that, in principle, at the stage of training, it is possible to introduce an additional component into the objective function to estimate real (financial) losses and gains of certain classifier decisions.

To improve the overall quality, we optimized the architecture of C1--C3 networks (the order, number and configuration of the layers, the number and configuration of the convolution kernels, etc.) applying the differential evolution method~\cite{bib:book_Price_2005}. To prevent overfitting, we used {\it dropout} and~$L_2$ regularization methods~\cite{bib:Book_Goodfellow-et-al_2016}.

After the primary training and optimization of the classifier, the adjustment of the training and test sets mark-up was carried out by the classifier itself. At this stage, we used rules~(\ref{eq:select_L2}) and~(\ref{eq:ThresCl}) to correct the vector for~$\mathbf{t}$ labels. Next, we performed the final tuning of the primary classifier. Note that we present the results of the t-SNE analysis (see Sect.~\ref{sec:contour}) and the classifier quality parameters (see the section below) at the stage of primary training, i.e., before the data set filtering and tuning of weights of the neural networks. We trained the networks for $\approx200$~epochs using the~SGD method~\cite{bib:Book_Goodfellow-et-al_2016}.

\subsection{The Results}
\label{sec:results}

To estimate the integral quality of the classifier operation on the test set, we have used standard Accuracy metric:
\begin{equation}
A = \frac{1}{K}\sum\limits^{K}_{k=1}
\begin{cases}
1 & \arg\max\limits_i \mathbf{t}_{k} = \arg\max\limits_i \mathbf{p}^\mathrm{ps}_{\mathrm{s}\,k},\\
0 & \text{otherwise}.\qquad i=0,\ldots,6
\end{cases}\:.
\end{equation}
The C1--C3 networks have achieved the following accuracies: $A^{(1)}\approx91.20$\%, $A^{(2)}\approx93.37$\%, and~$A^{(3)}\approx91.12$\%.

We have performed detailed investigation of the primary classifier accuracy and of its components based on the confusion matrix analysis and the density of classifier decisions distribution under various signal-jamming environment conditions. By way of example, the error matrix of the~C1 network is given in Table~\ref{tbl:ConfMatrix}.
\begin{table}[h!]
\begin{center}
\caption{Classifier C1: confusion matrix, test set.}
\label{tbl:ConfMatrix}
\small{
\begin{tabularx}{359pt}[c]{
|p{0.018\linewidth}
|p{0.018\linewidth}
|p{0.07\linewidth}
|p{0.07\linewidth}
|p{0.07\linewidth}
|p{0.07\linewidth}
|p{0.07\linewidth}
|p{0.07\linewidth}
|p{0.07\linewidth}|}\hline
\multicolumn{2}{|c|}{Event}&\multicolumn{7}{c|}{Prediction, \%}\\ \cline{3-9}
\multicolumn{2}{|c|}{class}
&$\phantom{00}0$&$\phantom{00}1$&$\phantom{00}2$&$\phantom{00}3$
&$\phantom{00}4$&$\phantom{00}5$&$\phantom{00}6$\\ \hline
\multirow{7}{*}{\begin{sideways}Reference, 100\%\end{sideways}}
&$0$&$91.80$&$\phantom{0}3.96$&$\phantom{0}0.64$&$\phantom{0}0.14$&$\phantom{0}2.34$&$\phantom{0}0.42$&$\phantom{0}0.70$\\\cline{2-9}
&$1$&$13.78$&$79.24$&$\phantom{0}5.38$&$\phantom{0}0.02$&$\phantom{0}1.14$&$\phantom{0}0.28$&$\phantom{0}0.16$\\\cline{2-9}
&$2$&$\phantom{0}4.24$&$\phantom{0}3.34$&$91.30$&$\phantom{0}0.14$&$\phantom{0}0.66$&$\phantom{0}0.00$&$\phantom{0}0.32$\\\cline{2-9}
&$3$&$\phantom{0}2.36$&$\phantom{0}0.10$&$\phantom{0}0.28$&$97.08$&$\phantom{0}0.12$&$\phantom{0}0.00$&$\phantom{0}0.06$\\\cline{2-9}
&$4$&$\phantom{0}8.68$&$\phantom{0}0.38$&$\phantom{0}0.22$&$\phantom{0}0.00$&$89.80$&$\phantom{0}0.28$&$\phantom{0}0.64$\\\cline{2-9}
&$5$&$\phantom{0}3.34$&$\phantom{0}0.14$&$\phantom{0}0.02$&$\phantom{0}0.00$&$\phantom{0}0.30$&$94.40$&$\phantom{0}1.80$\\\cline{2-9}
&$6$&$\phantom{0}3.42$&$\phantom{0}0.12$&$\phantom{0}0.14$&$\phantom{0}0.00$&$\phantom{0}0.70$&$\phantom{0}0.82$&$94.80$\\\hline
\multicolumn{2}{|c|}{Prec.} &$71.93$&$90.79$&$93.18$&$99.69$&$94.47$&$98.13$&$96.26$\\\hline
\multicolumn{2}{|c|}{F1 sc.}&$80.66$&$84.62$&$92.23$&$98.37$&$92.07$&$96.23$&$95.53$\\\hline
\end{tabularx}
}
\end{center}
\end{table}

\begin{figure}[htb]
\begin{minipage}[b]{1.0\linewidth}
\centerline{\includegraphics[width=8.5cm]{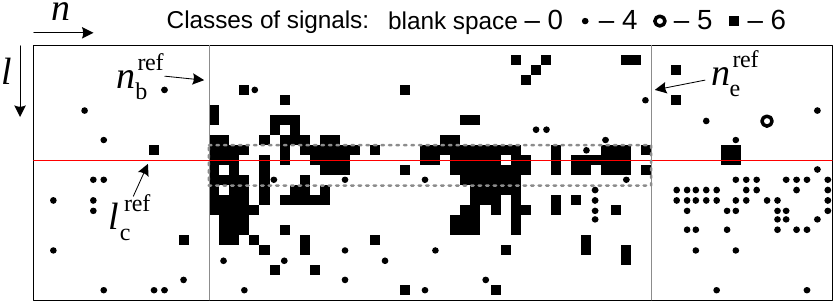}}
\end{minipage}
\caption{The decision of the~C1 network for validation sample~\#269 (class~6 signal). The mark-up denotes: $n^\mathrm{ref}_{\mathrm{b}}$ and~$n^\mathrm{ref}_{\mathrm{e}}$ are reference marks for signal beginning and end; $l^\mathrm{ref}_{\mathrm{c}}$ is reference (central) spatial channel; gray dotted rectangle is the signal-event track generated by the secondary classifier.}
\label{fig:Map_Cl_1_Sg_6_269}
\end{figure}

By way of example, Fig.~\ref{fig:Map_Cl_1_Sg_6_269} shows spatial-temporal representation of the~C1 network decision after application of rule~(\ref{eq:ThresCl}) for validation sample~\#269 (not included into the training set), which contains a class~6 signal.

The Fig.~\ref{fig:Map_Cl_1_Sg_6_269} indicates that the~C1 network has clearly and accurately identified the time-space frames that contain test signal of class~6, thus enabling the secondary classifier to generate an signal-event track. The parameters of the output track have turned out highly accurate with regard to the reference parameters, leading to a true-positive decision on detection of a class~6 event.

The level of error in the C1 network disproportionately leaning toward class~4 signals correlates with the data from t-SNE analysis (see Fig.~\ref{fig:tSNE_3D_Md_Ts_Set}) and can be explained by the signal portrait of this class being challenging to recognize (see Fig.~\ref{fig:PSD_Sg_JE_FrMn}). However, the ensemble of networks~C1-C3 suppresses these deviations quite effectively. Note that the probabilities of error in the primary classifier output cannot by any means be considered as final in terms of event decisions, as the system is equipped with the secondary classifier (see Sect.~\ref{sec:second}) that substantially decreases the levels of type~I and~II errors (as well as the ratio between them), when the decisions on frames are followed by decisions on signal-event tracks and the final event decision.

Finally, it is worth noting that, depending on the network, forward pass through~C1--C3 networks takes from~12 to~25~ms for~1000 distance channels computed on GPU NVIDIA GeForce GTX Titan~X. We see that the classifier facilitates real-time decisions for a monitored area of 50 km length, with spare hardware capacity that provides opportunity for further development of the algorithms.

\section{Conclusion}
\label{sec:conclusions}

In this paper, we have shown that efficient signal recognition algorithms for distributed fiber optic monitoring and security systems for long perimeters can be created based on deep learning methods. The integral averaged quality of a trained neural network of the primary multiclass detector is no less than~91\% of correctly recognized time-space data frames in the detector. We have obtained this estimate without applying the network ensemble or network weights tuning in conditions simulating actual operation conditions of the system to the maximum extent. Unlike other approaches, as for example~\cite{bib:book_Song_2002} or when only "hand-engineered" feature extraction technique is used, deep learning methods require less development effort and, most importantly, they allow for highly flexible reorganization of the system in case of changes in the classification of recognized events and/or in signal-jamming environment parameters.

It is worth noting that we are continuing to improve the parameters of the synthesized algorithms in different ways, among other things, by applying new deep architectures of neural networks.

Moreover, we have shown that nonlinear dimension reduction method t-SNE~\cite{bib:article_Maaten_JMLR_2008_9_2579} can be an efficient tool for preliminary analysis of informative features in terms of differentiation of signal classes. This allows to bypass the training of deep neural networks at the preliminary stage of synthesizing primary features, which saves time and hardware capacity.

In conclusion, it is worth noting that the demonstrated approach to developing signal recognition algorithms based on deep learning can be successfully extended to other applications, including radiolocation, hydroacoustics, supersonic or magnetic scanning of materials, etc. This conclusion is supported by the fact that the studied device (coherence optical time domain reflectometer) has nonlinear and nonstationary mode of operation~\cite{bib:article_Masoudi_RevSciInstr_2016_87_011501} and is used in difficult signal-jamming environments, whereas the algorithms based on convolutional neural networks counteract these negative factors quite successfully.

\section{Acknowledgments}

The authors thank the anonymous referees for their useful comments.


\bibliographystyle{IEEEbib}
\bibliography{strings4, refs}

\noindent
\\\textsf{\textbf{Andrey V. Makarenko} -- was born in~1977, since~2002 -- Ph.~D. of Cybernetics. Founder and leader of the Research \& Development group "Constructive Cybernetics". Author and coauthor of more than 60~scientific articles and reports. Member~IEEE (IEEE Signal Processing Society Membership; IEEE Computational Intelligence Society Membership). Research interests: Analysis of the structure dynamic processes, predictability; Detection, classification and diagnosis is not fully observed objects (patterns); Synchronization and self-organization in nonlinear and chaotic systems; System analysis and math.~modeling of economic, financial, social and bio-physical systems and processes; Convergence of Data~Science, Nonlinear~Dynamics, and~Network-Centric.}

\end{document}